# Analysis of Multiscale Wavelet-based Fractional Gradient-Anisotropic Diffusion Fusion for single hazy and underwater image enhancement

Uche A. Nnolim*

*Department of Electronic Engineering, Faculty of Engineering, University of Nigeria, Nsukka, Enugu, Nigeria*
*uche.nnolim@unn.edu.ng*

**Abstract:** *This report presents the results of a multi-scale wavelet based scheme for single image dehazing and underwater image enhancement. The scheme is fast and highly localized in addition to global enhancement of hazy images. A PDE-based formulation enables additional versatility as the iterative nature allows more flexibility for various types of images. Visual and objective results from experiments indicate that the proposed approach competes favourably or surpasses most of the state-of-the-art approaches.*

## 1. Introduction

The research area in image dehazing has grown tremendously over a relatively short time from enhancement and restoration models to fusion and highly involved machine and deep learning-based approaches. The subject of visibility restoration is a key aspect of intelligent transport systems, e.g. aerial unmanned systems and driverless cars among others. Over the years, the issues of over-enhancement and distortion in de-hazed images is still an issue, especially where the reference image scene is unavailable. Moreover, only a few algorithms can actually work across a relatively wide range of image datasets. Most algorithms may work well with some images and fail with others inspite of the training with available data. Also, this makes the deep learning approaches highly cumbersome and requiring a large amount of computational effort and execution in addition to data. In past works, PDE- and fusion-based approaches have been shown to be relatively low-cost alternative solution to the problem. Additionally, the parameters just require tuning for new sets of images, in similar ways as the deep learning approaches to achieve improved results.

Thus, the motivation for this work lies in the aim to develop an effective, reliable and relatively low complexity solution to the problem of single image de-hazing and underwater image enhancement.

The rest of the work is as follows; section 2 presents some background information on the various recent and some classical approaches in the literature on de-hazing and underwater image enhancement. Section 3 presents the proposed methods, while section 4 discusses results. The final section deals with the conclusion.

## 2. Background

Image de-hazing algorithms are usually classified under enhancement, restoration, fusion and machine or deep learning-based frameworks. However, recent works may combine two or more of the categories resulting in hybrid approaches.

Machine learning- and deep learning-based methods include those by Ancuti et al [1] [2], Galdran et al [3] [4], Singh and Biswas [5], and Nnolim [6]. The deep learning-based methods include those by Ren et al [7], Du and Li [8], Santra et al [9], Song et al [10], Zhang et al [11], J. Li et al [12], Lu et al [13], Ki et al [14], C. Li et al [15] and R. Li et al [16], Dehazenet [17], Uplavikar et al [18], Dudhane & Murala [19], Engin et al [20], B. Li et al, Mondal et al [21], Swami & Das [22], Zhang & Patel, et al [23] [11], Tang et al [24], Yan et al [25] and Yang et al [26]. The proposed scheme falls under enhancement and fusion-based frameworks and derives from the gradient domain fusion scheme by Paul et al [27] and anisotropic diffusion fusion by Bavirisetti and Dhuli [28].

## 3. Proposed algorithms

The proposed schemes are of two types; the first utilizes a two stage fusion approach by first performing a local contrast enhancement-based fusion in the gradient domain for both extracted multiscale base and detail layers of the image. The second stage performs a global and local enhancement-based fusion using the anisotropic diffusion fusion method by. The flowchart of the proposed algorithm is shown in Fig. 1. This is for effective image de-hazing any type of image.

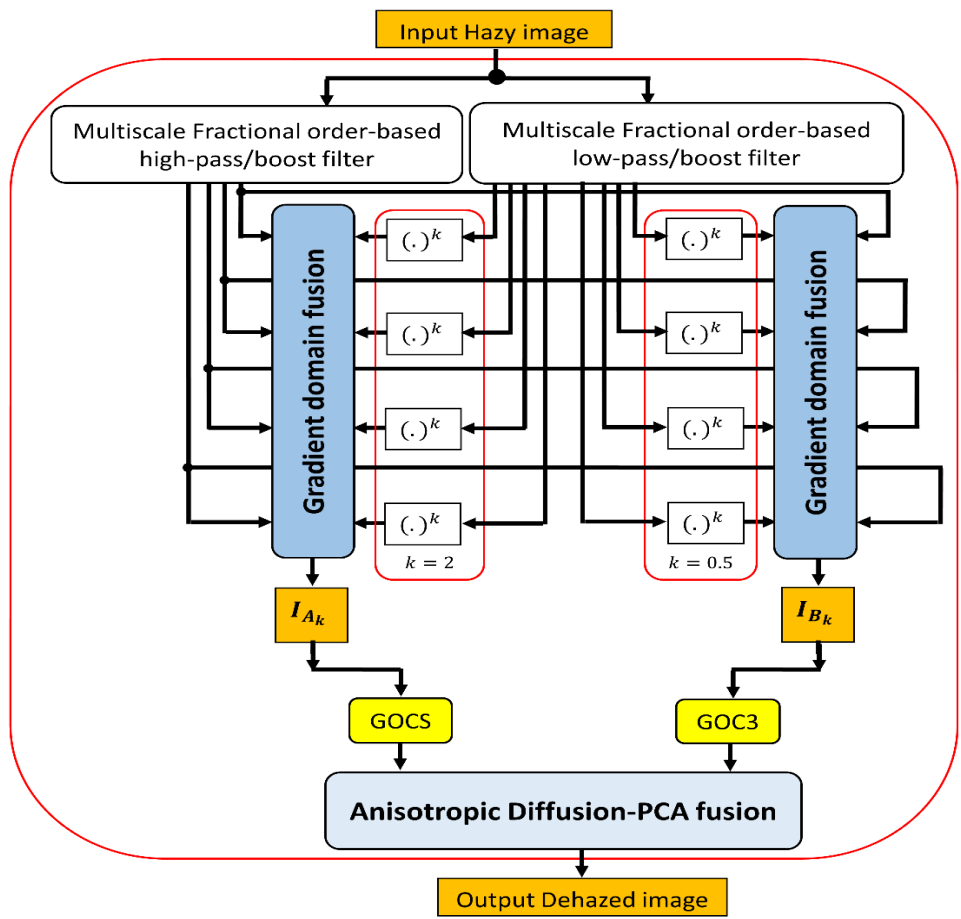

Figure. 1. Flowchart of PA1

For the second algorithm, we propose a wavelet-based approach to improve speed and localized contrast enhancement in addition to colour correction for underwater images. The use of the XYZ colour space is to minimize the effect of the green or blue colour dominance due to the depth at which the image was captured. The flowchart of the second proposed approach can be seen in Fig. 2.

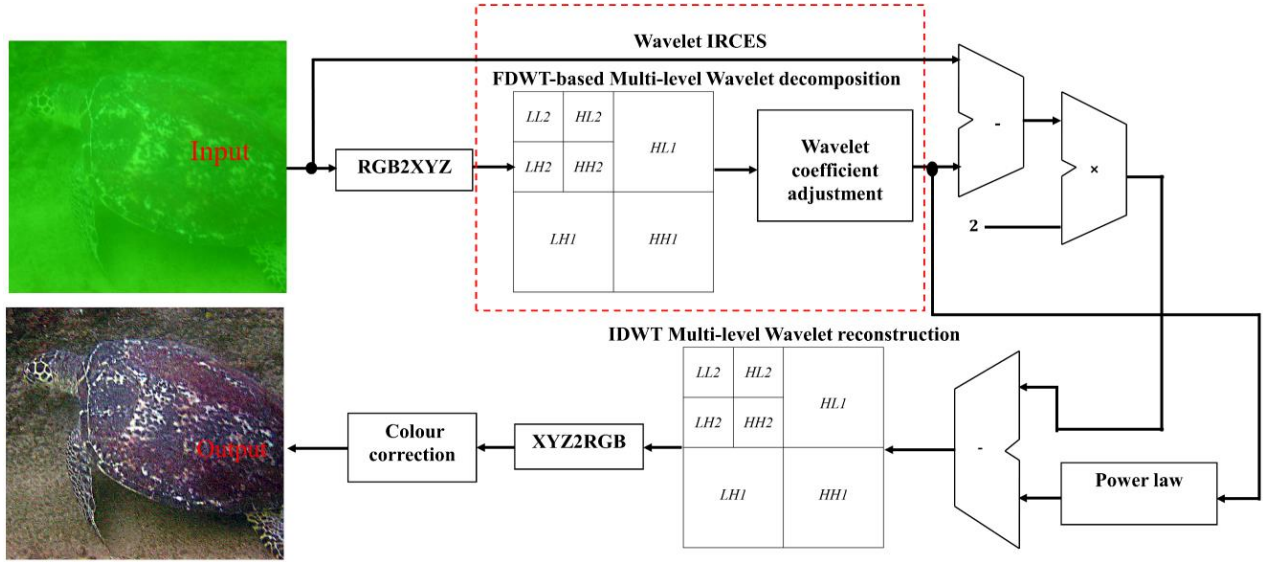

Figure. 2. Fractional Wavelet multi-level decomposition method for enhancing underwater images (PA2)

## 4. Experiments

We present some results from experiments involving the proposed approaches and their comparison with various methods from the literature **[29]**. These include He et al **[30]**, dehazing via adaptive dynamic stochastic resonance (DSR) **[31]**, PDE Retinex **[32]**, and other relevant dehazing and underwater image enhancement/restoration algorithms from the literature **[6] [29]**. We also use samples from the LIVE **[33]** and O-HAZE **[34]** hazy image datasets.

In Fig. 3, we present the enhancement results of PA-1 for underwater image frames from **[35]**. The pronounced contrast enhancement of PA-1 is clearly evident in the results. In Fig. 4, dehazing results for PA-2 are shown, depicting its persistent local contrast enhancement attributes. Even with thick haze, PA-2 improves on details.

In Fig. 5, we can see the effects of varying the parameters of PA-2 on the image results thus, by increasing the scales, we may get more localized enhancement but with increased halos and with fewer scales, there is more global enhancement and less or no halos. However, this can also cause darkening with less details as seen.

In Fig. 6, we compare against DCP, which darkens the images, over-enhances sky regions and produces excessive halos in addition to longer runtimes.

In Fig. 7, we compare against DSR, which darkens the image. In Fig. 8, we compare PA-1 and PA-2 where the former yields a darker image due to less versatility compared to the wavelet-based framework of PA-2.

In Fig. 9, PA-2 yields the brightest and more detailed de-hazed image with minimal halos compared to the other schemes. In Fig. 10, PA-2 yields the closest image to the ground truth image for the O-HAZE dataset. Overall, both PA-1 and PA-2 yield more consistently good results based on several experiments and at a fraction of the cost and time it takes for training large networks with massive amount of data and massive number of parameters and components. The proposed approaches can also be tuned to match the performance of most machine and deep learning-based methods with minimal effort since the schemes have much fewer parameters.

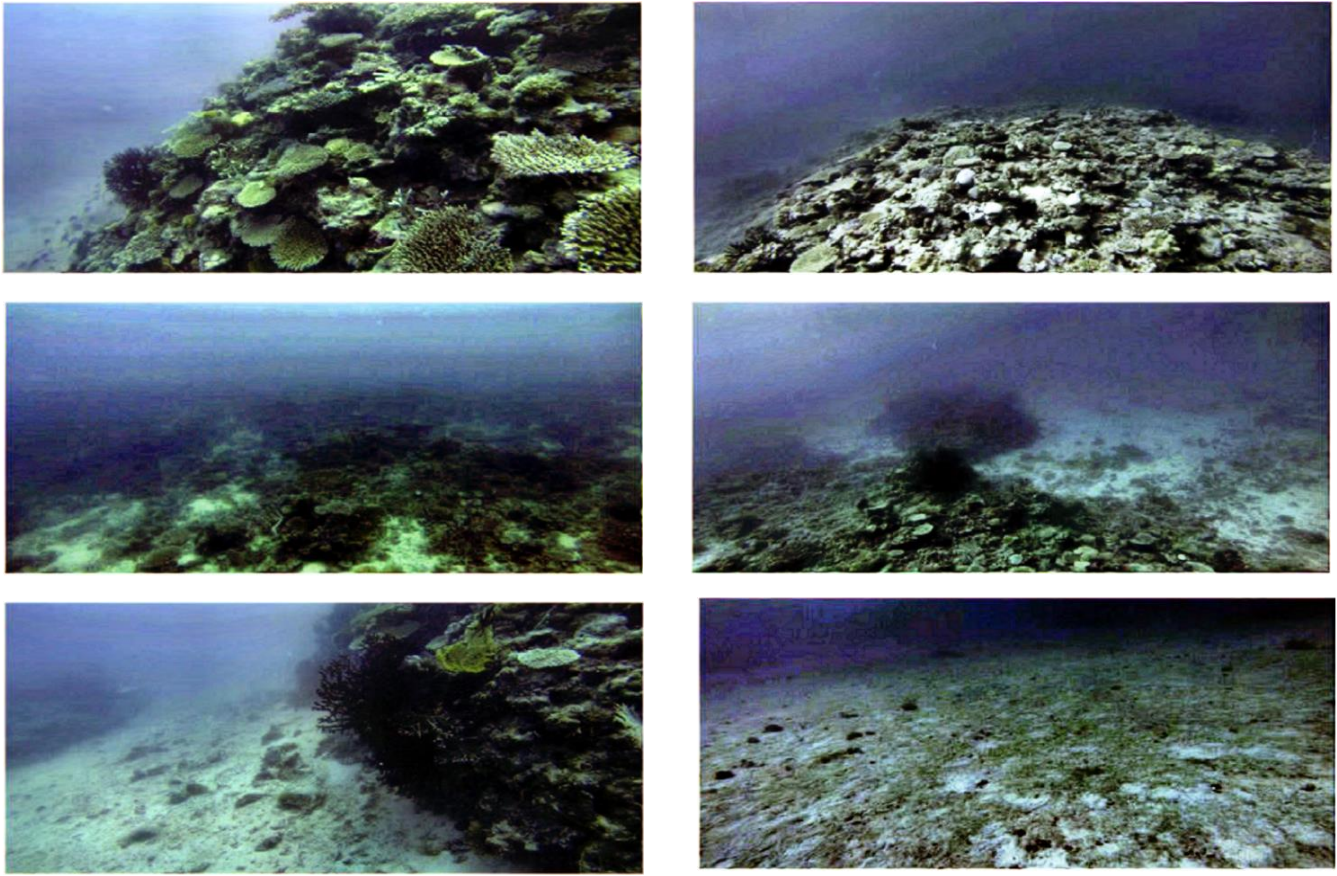

Fig. 3. Processed underwater video frames from [35] and images from using (b) PA-1

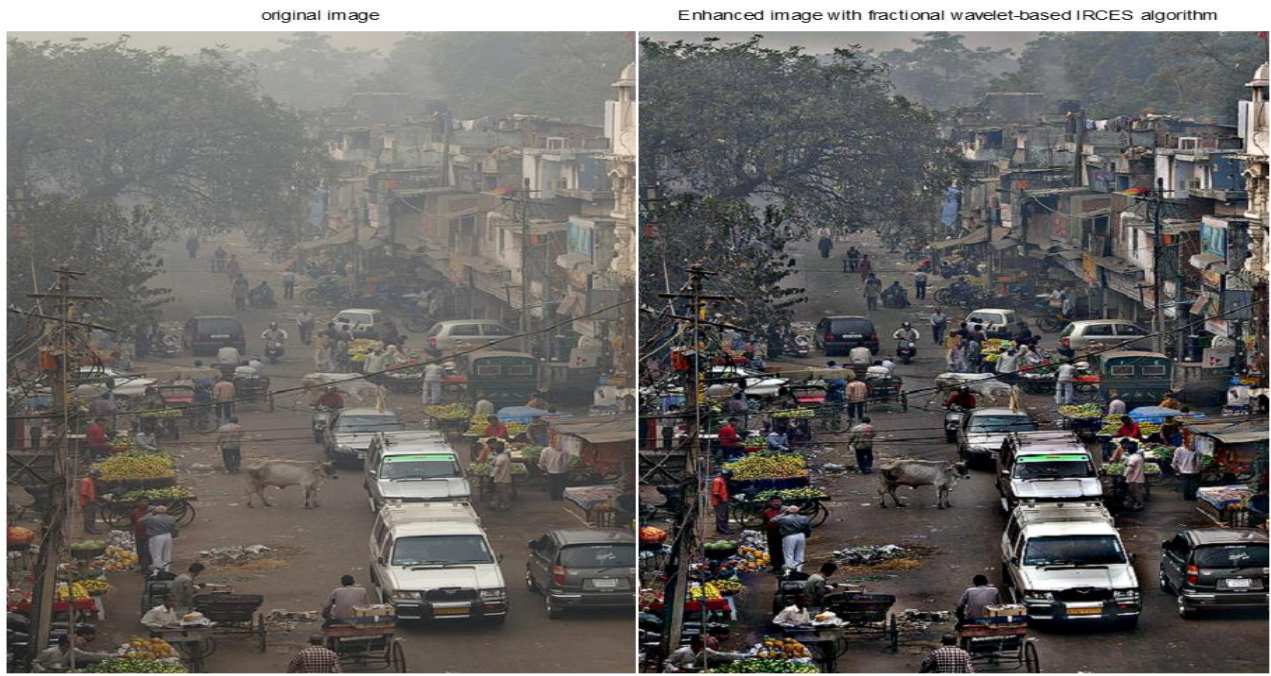

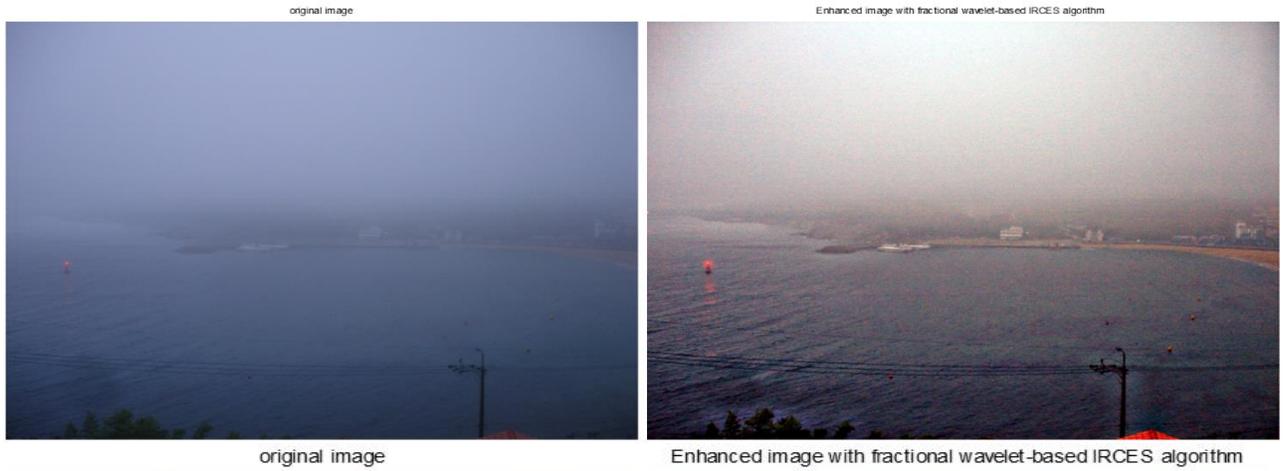

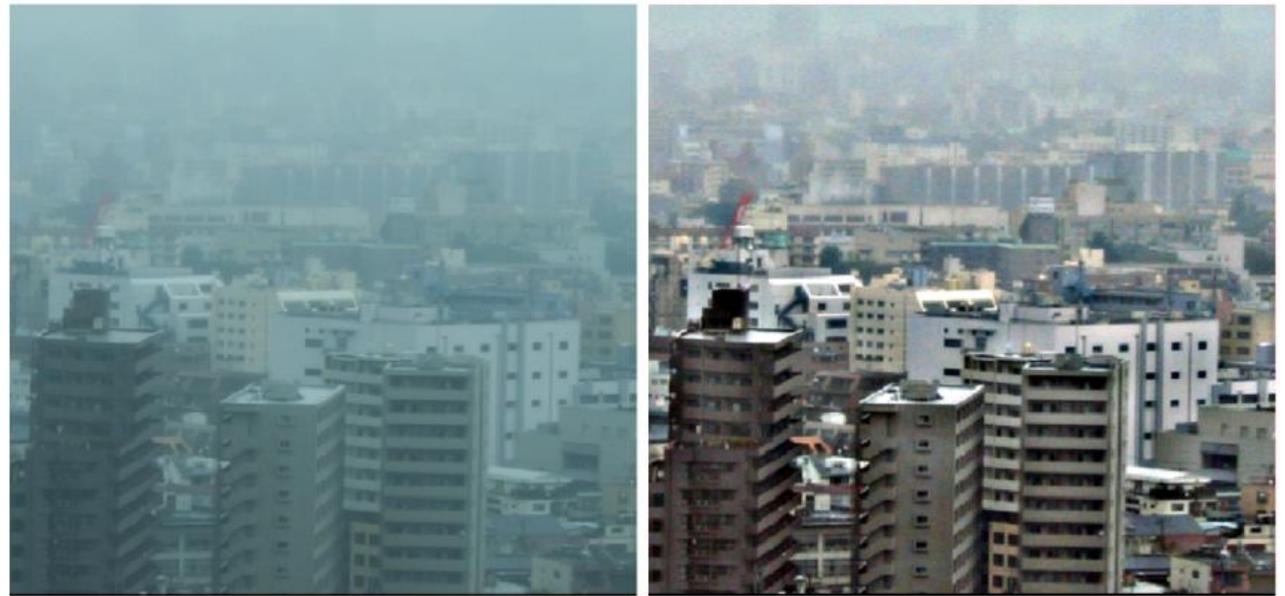

(a) (b)

Fig. 4. (a) Original hazy image and dehazed image using (b) PA-2

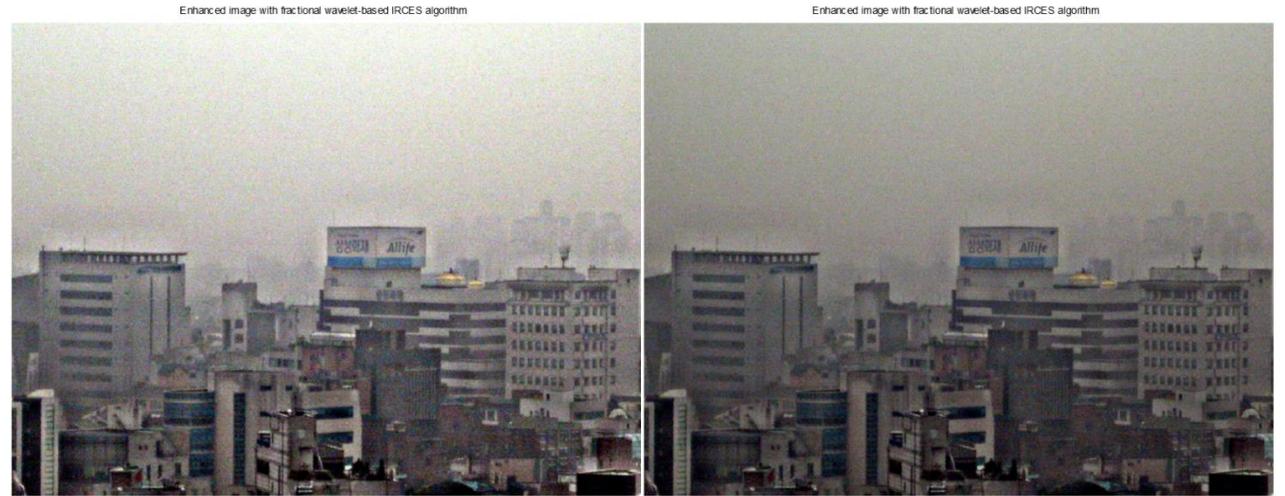

Fig. 5. Results with varying parameter settings of PA-2

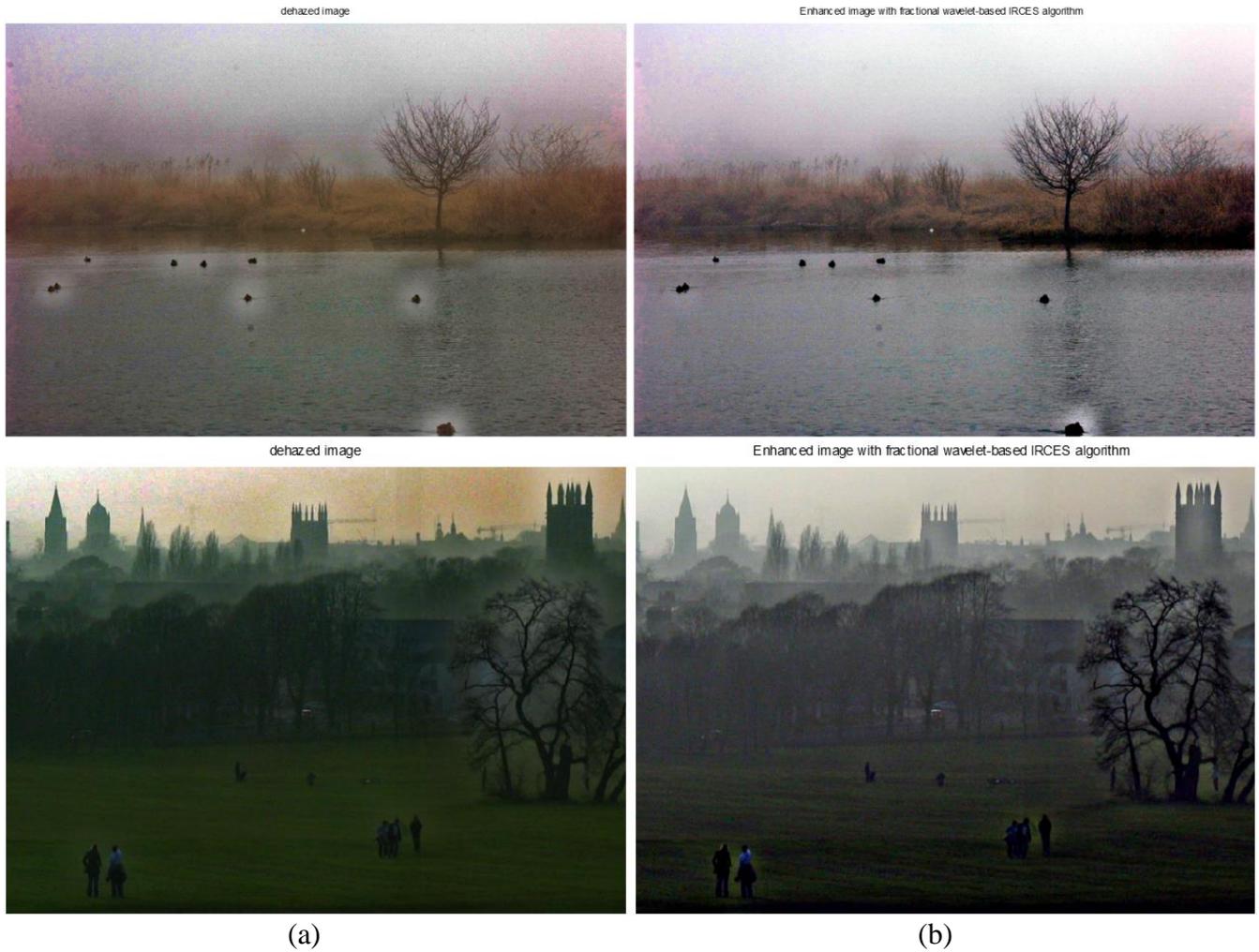

(a)                                              (b)

Fig. 6. Hazy image processed with (a) DCP by He et al and using (b) PA-2

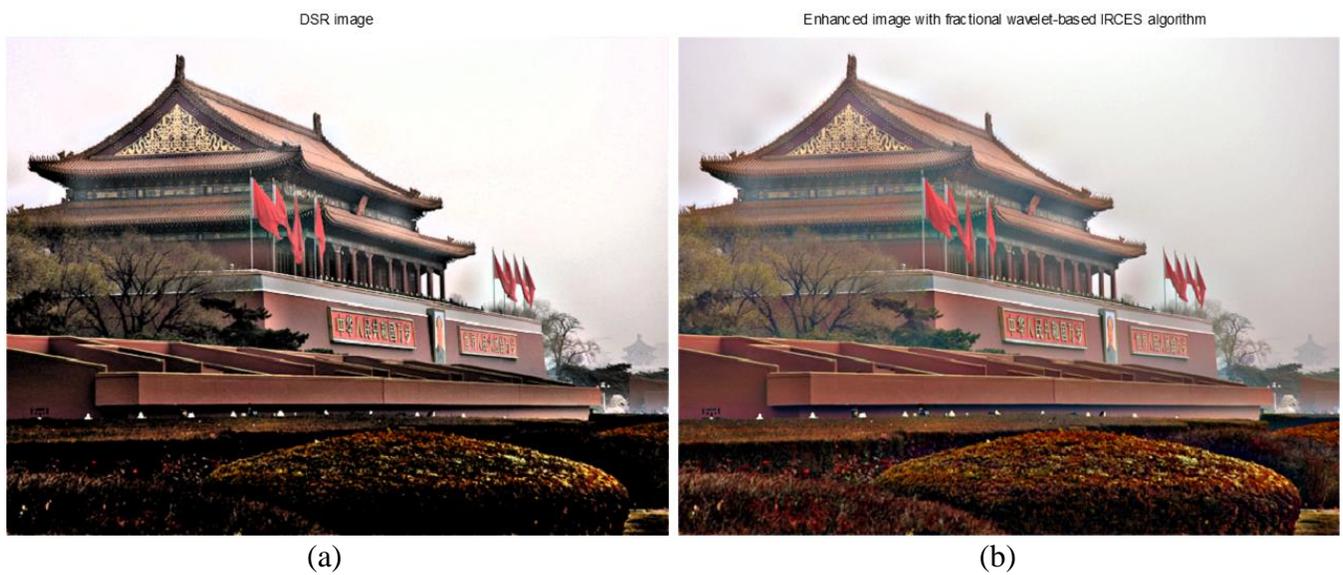

(a)                                              (b)

Fig. 7. Hazy image processed with (a) DSR and using (b) PA-2

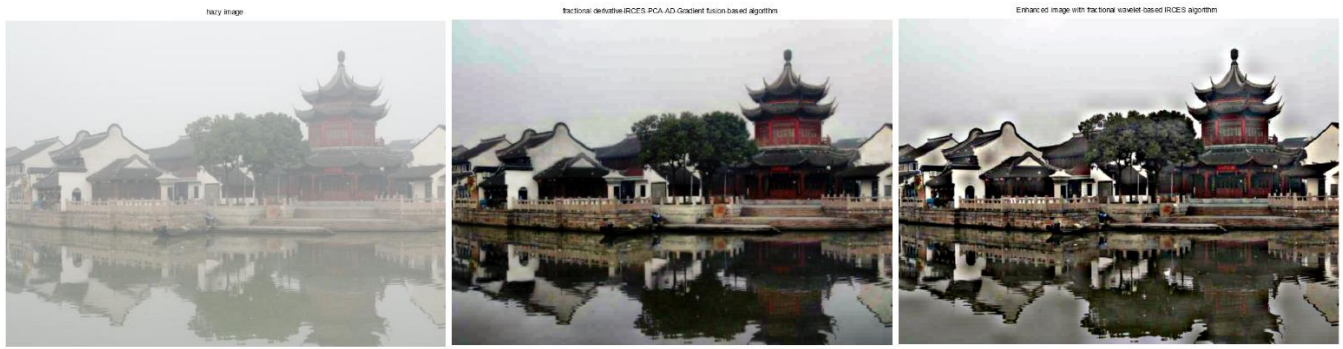

Fig. 8. (a) Original hazy image and dehazed image using (b) PA-1 and (c) PA-2

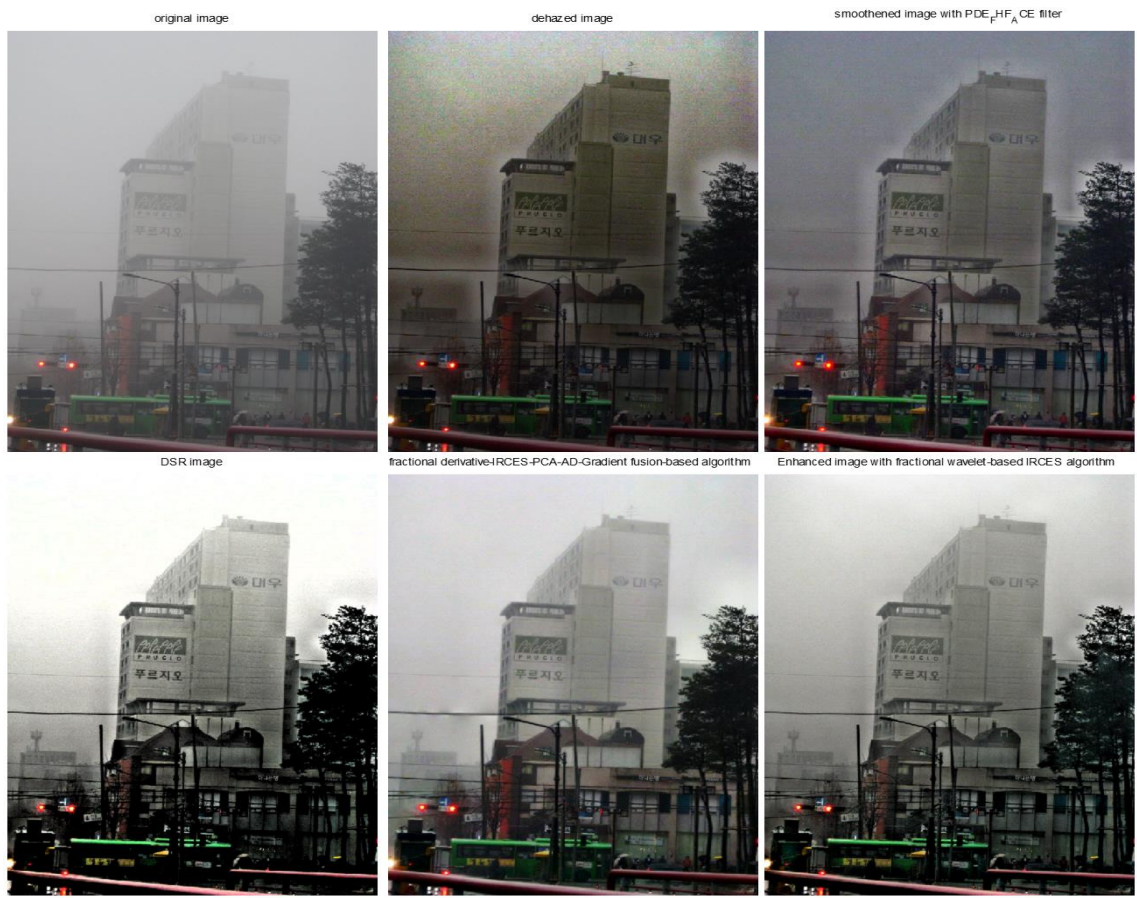

Fig. 9. (from left to right) (a) Original hazy image and (b) DCP by He etal (c) PDE-Retinex (d) DSR (e) PA-1 and (f) PA-2

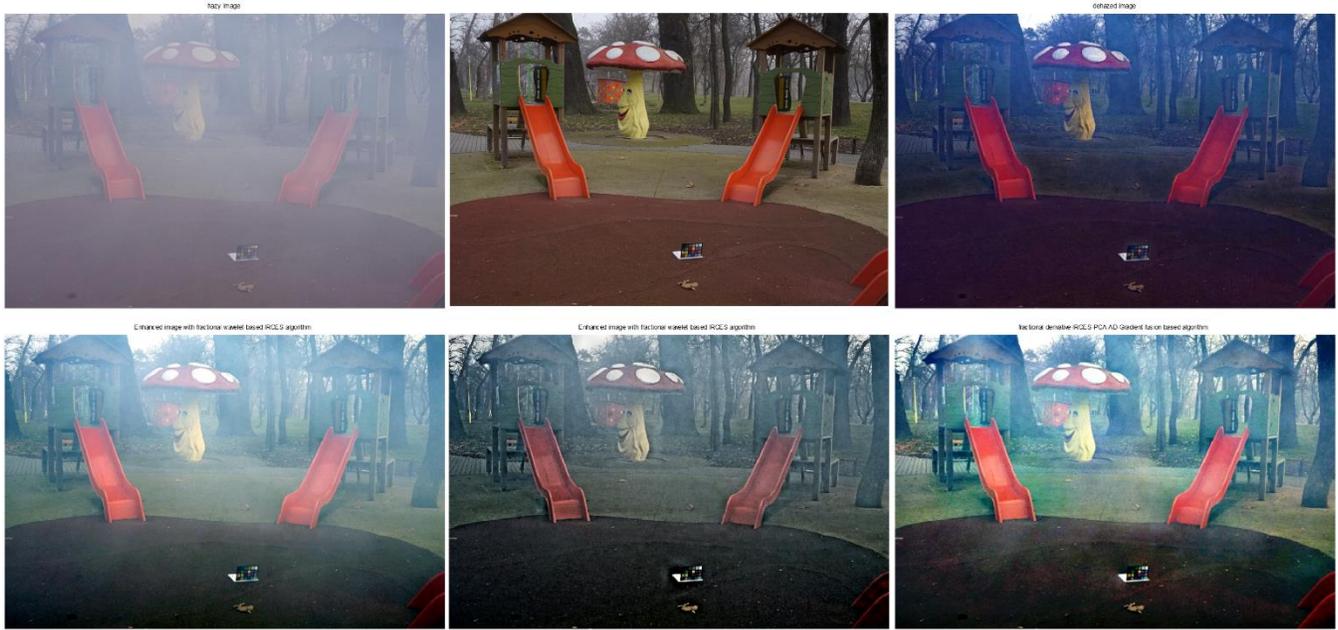

Fig. 10. (from left to right) (a) Original hazy image and (b) ground truth (c) DCP by He etal (d)-(e) PA-2 and (f) PA-1

## 5. Conclusion

We have presented the results of a proposed hazy and underwater image enhancement algorithms based on fusion and enhancement frameworks. The proposed schemes show good results and versatility in their ability to handle a wide variety of hazy and underwater images. They are also less computational complex than the deep learning-based approaches and do not require massive amount of computational and data resources nor are their execution times in anyway close to that of deep learning-based approaches.

## References

[1] Codruta Orniana Ancuti, Cosmin Ancuti, and Philippe Bekaert, "Effective single image dehazing by fusion," in *17th IEEE International Conference on Image Processing (ICIP)*, Hong Kong, China, 26–29 September 2010, pp. 3541-3544.